%% file: main.tex
\documentclass[runningheads]{llncs}

\usepackage[year=2026]{eccv} 
\usepackage{eccvabbrv}
\usepackage{graphicx}
\usepackage{booktabs}
\usepackage{amsmath}
\usepackage{amssymb}
\usepackage{xcolor}
\usepackage{multirow}
\usepackage{array}
\usepackage[accsupp]{axessibility}
\usepackage{hyperref}
\usepackage{orcidlink}
\usepackage[utf8]{inputenc}
\newcommand{\eh}[1]{\textcolor{black}{#1}}

\begin{document}

\title{Visual Words Meet BM25: Sparse Auto-Encoder Visual Word Scoring for Image Retrieval}

\titlerunning{BM25-V}

\author{
    Donghoon~Han \inst{1}\and 
    Eunhwan~Park \inst{2}\and
    Seunghyeon~Seo \inst{3}
} 
\authorrunning{Han et al.}
\institute{
    \mbox{
    \inst{1}D.notitia \quad
    \inst{2}Moreh \quad
    \inst{3}Seoul National University 
    }
}

\maketitle

% ---------------------------------------------------------------
\begin{abstract}

Dense image retrieval is accurate but offers limited interpretability and attribution, and it can be compute-intensive at scale. We present \textbf{BM25-V}, which applies Okapi BM25 scoring to sparse visual-word activations from a Sparse Auto-Encoder (SAE) on Vision Transformer patch features. Across a large gallery, visual-word document frequencies are highly imbalanced and follow a Zipfian-like distribution, making BM25's inverse document frequency (IDF) weighting well suited for suppressing ubiquitous, low-information words and emphasizing rare, discriminative ones. BM25-V retrieves high-recall candidates via sparse inverted-index operations and serves as an efficient first-stage retriever for dense reranking. Across seven benchmarks, BM25-V achieves Recall@200 $\geq$ 0.993, enabling a two-stage pipeline that reranks only $K{=}200$ candidates per query and recovers near-dense accuracy within $0.2$\% on average. An SAE trained once on ImageNet-1K transfers zero-shot to seven fine-grained benchmarks without fine-tuning, and BM25-V retrieval decisions are attributable to specific visual words with quantified IDF contributions.
\keywords{Image Retrieval \and Sparse Retrieval \and BM25 \and Sparse Auto-Encoder \and Interpretability}
\end{abstract}

% ---------------------------------------------------------------
\input{chapters/introduction}

\input{chapters/related_work}
\input{chapters/method}
\input{chapters/experiments}
\input{chapters/zipfian}
\input{chapters/conclusion}
% \input{chapters/supplement}

% ---------------------------------------------------------------
\bibliographystyle{splncs04}
\bibliography{main}

\end{document}

%% file: chapters/introduction.tex
\section{Introduction}
\label{sec:intro}
% are inherently lacked interpretabilty. For example, there is no answer to ``\textit{why was this retrieved``}}

% This paradigm is accurate, but carries two persistent limitations.
% First, dense embeddings are inherently \emph{opaque}: there is no answer to ``why was this image retrieved?''---a concern particularly in applications where retrieval decisions should be auditable, such as medical imaging, forensic search, and e-commerce.
% Second, full-precision dense indices require $O(N \cdot D)$ float32 memory
% ($N$ images, $D$-dimensional embeddings).
% At billion-scale, this is prohibitive.
% Product Quantization (PQ) saves memory by 8--32$\times$, but at the cost of recall degradation---a forced accuracy-memory tradeoff.

% \paragraph{What dense embeddings miss.}
% Dense retrieval pools all spatial information into a single Mean Average Pooled~(MAP) embedding.
% Fine-grained retrieval, however, often hinges on loca®l discriminative parts: a wing stripe, a wheel arch curvature, a petal arrangement.
% These part-level cues are compressed away by MAP pooling but preserved across patch tokens, where sparse encoding can amplify rare discriminative patches and suppress common background patches.

Image retrieval at scale is dominated by dense \eh{retrieval}: a vision encoder~\cite{radford2021learning,oquab2023dinov2,zhai2023sigmoid,tschannen2025siglip2} maps images to continuous embeddings, and retrieval is performed via Approximate Nearest Neighbor (ANN) search~\cite{johnson2021billion,malkov2018efficient}.
Since this paradigm is simple yet effective, it remains several drawbacks. First, dense retrieval offers limited interpretability and attribution, thereby raising the concerns in real-world applications where decisions should be auditable, such as medical imaging, forensic search, and e-commerce, respectively. Second, since retrieving full-precision indices require $O(N \cdot D)$ float32 memory where $N, D$ indicate images and dimension, this is not practical for real-world application when billion-scale. Product Quantization (PQ)~\cite{pq} mitigates the usage of memory by $8 \sim 32\times$, however it has inevitably accuracy -- memory trade-off. Lastly, dense retrieval tends to lose fine-grained spatial evidence because it typically aggregates patch-level representations into a single global embedding via pooling. While this global summarization is effective for coarse semantics, it can suppress local discriminative evidence that are critical for fine-grained retrieval, such as subtle part shapes, textures, and localized patterns.

To recover \textit{interpretable} and \textit{index-friendly} evidence without discarding patch-level information, we revisit a classic sparse retrieval by represent each image as a small set of discriminative ``terms'' and retrieve with an inverted index. Concretely, inspired by findings of transformer representations that reveal monosemantic features in language models, we apply a SAE to late-layer ViT patch features and obtain sparse ``visual words'' activations~\cite{bricken2023monosemanticity}. We observe that SAE visual words exhibit the document-frequency imbalance across the gallery, consistent with a Zipfian-like distribution~\cite{zipf1949human}.
Rare words tend to capture discriminative visual evidence, while frequent words largely reflect generic content such as backgrounds and common textures.

This observation implies that a heavy-tailed term distribution is aligned to Okapi BM25~\cite{robertson1994okapi}, which is developed for sparse retrieval under document-frequency imbalance. In particular, BM25 employs inverse document frequency weighting that attenuates ubiquitous terms and emphasizes rare, discriminative ones. Motivated by this alignment, we adopt SAE-derived visual words with BM25 and refer to the resulting sparse retriever as \textbf{BM25-V}.

BM25-V enables efficient inverted-index retrieval by representing each image as a compact set of active visual words. Given late-layer ViT patch features, we apply the SAE encoder and aggregate activations across patches to obtain image-level visual-word weights, which we treat as term frequencies. We then retain only the top-$k$ words per image to form a sparse bag-of-visual-words representation. At query time, scoring is restricted to the term-specific inverted lists of the active query words, yielding per-query complexity on the order of $O(k \cdot \overline{\mathrm{df}})$ rather than the $O(N \cdot D)$ dense similarity computation required by full-scan retrieval.

In our experiments, BM25-V achieves Recall@100 comparable to dense Recall@10 across all seven benchmarks. This motivates a two-stage pipeline where BM25-V serves as a fast first-stage retriever and dense reranking is applied only to the top-$K$ retrieved candidates. As a result, dense similarity scoring is performed for only $K$ candidates per query rather than the full gallery, while maintaining near-dense accuracy within $0.2$ percentage points on average. In contrast, product quantization (PQ) reduces memory by compressing dense embeddings and can incur non-trivial accuracy degradation.
Our two-stage design instead preserves dense representations and reduces compute through candidate pruning. Moreover, the rerank stage stores dense embeddings at $4D$ bytes per vector in float32, and BM25-V adds a lightweight sparse index of 96 bytes per image when $k{=}16$ visual words are stored using a 4-byte id and a 2-byte activation. Although the combined memory footprint is higher than PQ (24--48 bytes per vector in our experiments), it avoids the 1--6\% accuracy degradation observed with PQ.

Our contributions are summarized as follows:
\begin{enumerate}
  \item \textbf{BM25-V: first application of Okapi BM25 to SAE-derived sparse visual words.}
  We introduce BM25-V, which applies Okapi BM25 to SAE-derived visual words for image retrieval. We show that the resulting visual-word document frequencies are heavy-tailed (Zipfian-like), making BM25 a principled choice in this sparse feature space. We further validate key design choices, including sparsity $k$ and expansion factor $e$, via systematic ablations (\cref{tab:k} and Appendix).

  \item \textbf{Two-stage retrieval with substantial compute reduction.}
  BM25-V provides a high-recall candidate set (e.g., R@100 comparable to dense R@10; R@200 $\geq$ 0.993 across seven benchmarks),
  and dense reranking over $K$ candidates recovers near-dense rank-1 accuracy (within $0.2$\% on average). This reduces dense similarity evaluations from $N$ gallery items to $K$ candidates per query.

  \item \textbf{Zero-shot cross-domain generalization.}
  An SAE trained once on ImageNet-1K~\cite{deng2009imagenet} transfers to seven fine-grained retrieval benchmarks without fine-tuning, suggesting that the learned visual vocabulary generalizes beyond the training distribution.

  \item \textbf{Interpretability by construction.}
  BM25-V retrieval decisions are attributable to discrete visual words with explicit IDF contributions, enabling transparent, term-level explanations (see supplementary material).
\end{enumerate}

%% file: chapters/related_work.tex
\section{Related Work}
\label{sec:related}
% \paragraph{Dense visual retrieval.}
% The standard image retrieval pipeline encodes images with a vision backbone~\cite{radford2021learning,oquab2023dinov2,zhai2023sigmoid,tschannen2025siglip2} and searches a MAP-pooled embedding using ANN indices~\cite{johnson2021billion,malkov2018efficient}.
% FAISS-HNSW~\cite{malkov2018efficient} achieves near-lossless approximate search; FAISS-IVF+PQ~\cite{johnson2021billion} reduces memory via Product Quantization at a recall cost.
% These systems are opaque by construction---dense embeddings distribute information across hundreds of entangled dimensions with no human-interpretable structure.
% BM25-V adds a complementary sparse channel built on the same backbone: by generating a high-recall first-stage candidate set via sparse scoring, it reduces the number of dense dot-products required while preserving near-exact accuracy---unlike PQ which trades accuracy for memory. Additionally, sparse inverted indices support truly dynamic gallery updates---adding or removing an image costs O(k) index operations---whereas HNSW's graph structure degrades with large-scale insertions and deletions require a full rebuild.
\paragraph{Dense visual retrieval.}
Standard image retrieval encodes images with a vision backbone~\cite{radford2021learning,oquab2023dinov2,zhai2023sigmoid,tschannen2025siglip2} and performs ANN search over MAP-pooled embeddings~\cite{johnson2021billion,malkov2018efficient}.
FAISS-HNSW~\cite{malkov2018efficient} provides near-lossless approximate search, while FAISS-IVF+PQ~\cite{johnson2021billion} reduces memory via product quantization at the cost of recall.
However, dense embeddings offer limited interpretability because retrieval evidence is distributed across many entangled dimensions.
BM25-V complements dense retrieval with a sparse channel that produces a high-recall candidate set for dense reranking, reducing dense similarity evaluations while preserving near-dense accuracy.
Inverted-index retrieval also supports efficient dynamic gallery updates, where inserting or deleting an image requires updating only its sparse entries.

\paragraph{Sparse text retrieval and BM25.}
Okapi BM25~\cite{robertson1994okapi}, rooted in the Probabilistic Relevance Framework~\cite{robertson1976relevance}, remains a strong zero-shot baseline for text IR.
Its effectiveness is closely tied to long-tailed term statistics, where inverse document frequency (IDF) downweights ubiquitous terms and emphasizes rare, discriminative ones.
Recent learned sparse retrievers such as DeepCT~\cite{dai2020deepct}, doc2query~\cite{nogueira2019doc2query}, uniCOIL~\cite{lin2021few}, and SPLADE~\cite{formal2021splade} preserve the inverted-index paradigm while learning term weights, and SPLADE in particular produces sparse activations that are naturally scored with BM25-style weighting.
We draw a direct parallel in vision.
SAE activations on ViT patch features exhibit a similar document-frequency imbalance, making BM25 a principled scoring choice for suppressing pervasive visual words and highlighting rare evidence.
Consistent with probabilistic relevance theory~\cite{robertson1976relevance}, frequent visual words carry limited discriminative power, and retrieval quality degrades without their suppression.

\paragraph{Hybrid sparse-dense retrieval.}
In text IR, combining sparse BM25 and dense neural retrieval~\cite{karpukhin2020dense} consistently outperforms either method alone.
The complementarity is well-established: BM25 captures exact lexical patterns at low cost, while dense retrieval captures semantic paraphrase.
Hybrid text systems have become standard practice in retrieval-augmented generation (RAG) pipelines.
We demonstrate an analogous complementarity in vision: BM25-V captures local discriminative part patterns (patch-level, IDF-weighted) while dense MAP embedding captures global class semantics.
The key non-obvious step is establishing that SAE visual words exhibit the same Zipfian distributional property that makes IDF weighting effective in text---this has not been shown before for visual features, and is not self-evident from either the SAE or ViT literature.
The pipeline architecture is therefore a known pattern; the contribution is the empirical demonstration that the prerequisite distributional condition holds in vision.

\paragraph{SAE-based retrieval.}
CL-SR~\cite{park2025decoding} applies SAEs to dense text retrieval embeddings to enable interpretable concept-level retrieval, scoring SAE activations with dot product.
SPLARE~\cite{formal2026learning} jointly train a retrieval model with SAE objectives for text.
We extend the SAE-retrieval paradigm to vision and observe that pervasive, uninformative dimensions must be suppressed for effective sparse retrieval.
BM25's IDF weighting is the principled mechanism for this suppression; our post-pool top-$k_{\text{post}}$ filter provides complementary hard filtering at inference.

\paragraph{Classical visual words.}
Bag-of-Visual-Words (BoVW)~\cite{sivic2003video} and VLAD~\cite{jegou2010aggregating} clustered hand-crafted descriptors (SIFT, HOG) into discrete visual words and scored with TF-IDF.
These methods were displaced by deep features.
BM25-V revives the visual word paradigm using SAEs trained on deep ViT features, obtaining monosemantic, semantically grounded visual words rather than the k-means cluster centroids used by BoVW and VLAD.
The inverted index efficiency is preserved, while the visual vocabulary is now learned end-to-end from deep representations.

%% file: chapters/method.tex
\section{Method}
\label{sec:method}

\subsection{Background: Okapi BM25}
\label{sec:prelim:bm25}

Given a corpus of $N$ documents and a query $Q$, Okapi BM25~\cite{robertson1994okapi} scores document $d$ as:
\begin{equation}
  \text{BM25}(d, Q) = \sum_{i \in Q} \text{IDF}_i \cdot \frac{f_i(d)\,(k_1 + 1)}{f_i(d) + k_1\!\left(1 - b + b\,\dfrac{|d|}{\overline{dl}}\right)},
  \label{eq:bm25}
\end{equation}
where $f_i(d)$ is term frequency, $|d|$ document length, $\overline{dl}$ mean document length, $k_1{=}1.5$, $b{=}0.75$, and:
\begin{equation}
  \text{IDF}_i = \ln\!\left(1 + \frac{N - \text{df}_i + 0.5}{\text{df}_i + 0.5}\right),
  \label{eq:idf}
\end{equation}
with $\text{df}_i$ the number of documents containing term $i$.
IDF assigns near-zero weight to pervasive terms and high weight to rare, discriminative ones---the decisive property when vocabulary usage is heavy-tailed.

\cref{tab:analogy} formalizes the correspondence:
sum-pooling SAE activations across patches is mathematically equivalent to counting term occurrences across document tokens, so all BM25 theory transfers directly.

\begin{table}[tb]
  \caption{The text--vision BM25 analogy is structural, not metaphorical.
           Sum-pooling SAE activations across all $P$ patch tokens is mathematically equivalent to
           counting term occurrences across document tokens.
           Here $\tilde{H} \in \mathbb{R}^{P \times eD}$ denotes the SAE activation matrix
           ($\tilde{H}[p,i]$ is the activation of dimension $i$ at patch $p$),
           and $eD$ is the SAE feature dimension.}
  \label{tab:analogy}
  \centering
  \small
  \begin{tabular}{@{}l|l@{}}
    \toprule
    \textbf{Text BM25} & \textbf{BM25-V (visual)} \\
    \midrule
    Document $d$ & Image $I$ \\
    Term vocabulary $\{t\}$ & SAE dimensions $\{0,\ldots,eD{-}1\}$ \\
    Term frequency $f_t(d)$ & $\mathbf{v}[i] = \sum_p \tilde{H}[p,i]$ \; (sum-pool) \\
    Document length $|d|$ & $\|\mathbf{v}\|_1$ \; (L1 norm) \\
    Document freq $\text{df}_t$ & $\text{df}_i = |\{j : \mathbf{v}_j[i] > 0\}|$ \\
    $\text{IDF}_t$ & $\text{IDF}_i = \ln\!\left(1 + \dfrac{N - \text{df}_i + 0.5}{\text{df}_i + 0.5}\right)$ \\
    Query presence & $\mathbf{v}_q[i] > 0$ \\
    \bottomrule
  \end{tabular}
\end{table}

\subsection{System Overview}
\label{sec:method:overview}

In this paper, \textbf{BM25-V} refers specifically to the sparse retrieval component
(first stage); the full system is called the \textbf{two-stage BM25-V pipeline}
(BM25-V first stage followed by dense reranking).
The two-stage pipeline operates over a frozen SigLIP2 backbone:

\begin{enumerate}
  \item \textbf{First stage (BM25-V).} Patch features are encoded into sparse visual word
  vectors and indexed with BM25. At query time, BM25-V scores all reference set images via
  sparse matrix operations and returns the top-$K$ candidates.
  \item \textbf{Second stage (Dense rerank).} The MAP pooler embedding of the query is
  compared against the $K$ candidate embeddings via cosine similarity, producing the
  final ranked list.
\end{enumerate}

A single backbone forward pass yields both the patch features (for BM25-V) and the
MAP pooler embedding (for reranking) simultaneously.

\subsection{Visual Word Extraction: Sparse Autoencoder}
\label{sec:method:sae}

We use SigLIP2-SO400M~\cite{tschannen2025siglip2} (\texttt{google/siglip2-so400m-patch14-384})
as our frozen visual backbone, extracting patch features
$\mathbf{Z}^{(\ell)} \in \mathbb{R}^{P \times D}$ at layer $\ell{=}26$
($P{=}729$ patches for this architecture, $D{=}1152$ feature dimensions).
A Sparse Autoencoder (SAE)~\cite{bricken2023monosemanticity} with expansion factor $e$
maps each patch feature to a sparse latent of dimension $eD$:
\begin{align}
  \mathbf{h} &= \operatorname{top-}k\!\left(\text{ReLU}\!\left(\mathbf{W}_e\,\mathbf{z}+\mathbf{b}_e\right)\right),
  \quad \|\mathbf{h}\|_0 = k, \label{eq:sae_enc}\\
  \hat{\mathbf{z}} &= \mathbf{W}_d\,\mathbf{h}. \label{eq:sae_dec}
\end{align}
Here $\operatorname{top-}k(\cdot)$ retains the $k$ largest activations by magnitude and zeros the rest;
$\mathbf{W}_d \in \mathbb{R}^{D \times eD}$ is the decoder weight matrix and $\hat{\mathbf{z}}$ the reconstruction.
The $eD$ dimensions of $\mathbf{h}$ are \emph{visual words}---monosemantic feature
directions~\cite{bricken2023monosemanticity}.
Training minimizes $\mathcal{L} = \|\hat{\mathbf{z}}-\mathbf{z}\|_2^2 + \lambda\|\mathbf{h}\|_1$.
Gradients flow through the $k$ retained activations unchanged and are zero for the masked entries (straight-through on the kept values); no explicit relaxation is needed.

\subsection{BM25-V: Sparse Retrieval Channel}
\label{sec:method:bm25v}

\paragraph{Patch feature extraction.}
Given image $\mathbf{I}$, we extract patch features at the final transformer layer ($\ell{=}26$):
\begin{equation}
  \mathbf{Z} = \mathbf{Z}^{(26)} \in \mathbb{R}^{P \times D},
  \label{eq:patchfeat}
\end{equation}
where $P$ is the number of patch tokens (architecture-dependent; $P{=}729$ for our backbone)
and $D{=}1152$ is the feature dimension.
Unlike dense retrieval, which discards patch structure via MAP pooling, we retain all $P$ patch tokens.
Ablation on our SigLIP2 backbone confirms that the final layer ($\ell{=}26$) is most discriminative for retrieval (\cref{sec:exp:ablation}).

\paragraph{SAE encoding.}
Each patch vector $\mathbf{z}_p \in \mathbb{R}^D$ is independently encoded by the SAE (\cref{sec:method:sae}):
% \[
\begin{equation}
  \mathbf{h}_p = \operatorname{top-}k\!\left(\text{ReLU}\!\left(\mathbf{W}_e\,\mathbf{z}_p + \mathbf{b}_e\right)\right) \in \mathbb{R}^{eD},
  \quad \|\mathbf{h}_p\|_0 = k.
\end{equation}
% \]
Each $\mathbf{h}_p$ is sparse with exactly $k$ non-zero entries---the $k$ visual words most active at patch location $p$.

\paragraph{Sum pooling as term frequency.}
We aggregate patch-level sparse vectors by sum pooling:
\begin{equation}
  \mathbf{v}_{\text{pool}} = \sum_{p=1}^{P} \mathbf{h}_p \in \mathbb{R}^{eD}.
  \label{eq:sumpool}
\end{equation}
The $i$-th dimension of $\mathbf{v}_{\text{pool}}$ records the total activation magnitude of visual word $i$ across all patches of the image.
This is the natural \emph{term frequency} analog: it counts how strongly and frequently each visual word appeared in the image, in the same way TF counts term occurrences across document positions.

\paragraph{Post-pool top-$k_{\text{post}}$ filter.}
A second top-$k_{\text{post}}$ is applied at the image level after sum pooling:
\begin{equation}
  \mathbf{v} = \operatorname{top-}k_{\text{post}}(\mathbf{v}_{\text{pool}}) \in \mathbb{R}^{eD}, \quad \|\mathbf{v}\|_0 = k_{\text{post}}.
  \label{eq:pptk}
\end{equation}
The SAE is trained with patch-level top-$k$ only; this filter is applied at inference.
It retains the $k_{\text{post}}$ dimensions with the highest accumulated activation---the image's dominant visual concepts---and discards the noisy long tail.

After sum pooling, empirical sparsity is ${\approx}17\%$ (${\approx}3{,}133$ active dims); post-pool top-$k_{\text{post}}$ eliminates noise tail dimensions that fired by chance across few patches (derivation in the supplementary material).

\paragraph{Quantization.}
$\mathbf{v}$ is stored as \texttt{uint16} (values $\times 100$, rounded, and clipped to $[0, 65535]$) and dequantized to float32 ($\div 100$) before BM25 scoring so that TF saturation operates on the correct scale.
This halves index memory and discretizes TF values to steps of $0.01$, with negligible impact on retrieval accuracy.

\paragraph{Reference set indexing and IDF computation.}
Given $N$ reference set images with visual word vectors $\{\mathbf{v}_1, \ldots, \mathbf{v}_N\}$, we compute:
\begin{equation}
  \text{df}_i = \left|\left\{j : \mathbf{v}_j[i] > 0\right\}\right|, \qquad
  \text{IDF}_i = \ln\!\left(1 + \frac{N - \text{df}_i + 0.5}{\text{df}_i + 0.5}\right).
  \label{eq:idf_df}
\end{equation}
IDF is computed once from the reference set (train split) and stored.
Dimensions that activate in nearly all images---common textures, lighting gradients, background patterns---receive IDF $\approx 0$.
Rare, semantically specific visual words receive high IDF.
This is precisely the mechanism by which BM25 suppresses uninformative visual words at scoring time.

\paragraph{BM25 scoring.}
BM25 scores query $\mathbf{v}_q$ against reference image $\mathbf{v}_d$ via \cref{eq:bm25},
setting $f_i(d){=}\mathbf{v}_d[i]$, $|d|{=}\|\mathbf{v}_d\|_1$, and activating dimension $i$
iff $\mathbf{v}_q[i]{>}0$.
The BM25-weighted index is pre-computed; query scoring reduces to a sparse matrix-vector
multiply in $O(k \cdot \overline{\text{df}})$ operations.

\subsection{Two-Stage Query Pipeline}
\label{sec:method:twostage}

Following the overview in \cref{sec:method:overview}, the first stage scores all $N$
reference images via BM25 in $O(k \cdot \overline{\text{df}})$ operations per query
and returns the top-$K$ candidates; the second stage reranks these $K \ll N$ candidates
by dense cosine similarity, reducing dense computation from $N{\cdot}D$ to $K{\cdot}D$.
BM25-V R@100 reaches $0.984$--$0.999$ across all seven benchmarks, so the rerank
stage operates on a near-complete recall pool (\cref{sec:exp:main}).

\subsection{SAE Training}
\label{sec:method:training}

The SAE is trained on ImageNet-1K patch features~\cite{deng2009imagenet} (5 epochs, batch 4096,
Adam with cosine LR decay, $\lambda{=}10^{-3}$), with SAE sparsity $k{=}16$ (non-zero activations per patch)
and expansion factor $e{=}16$ (vocabulary size $eD{=}18{,}432$).
Post-pool top-$k_{\text{post}}{=}16$ and quantization are inference-only; SAE training is unaware of them.

%% file: chapters/experiments.tex
\section{Experiments}
\label{sec:exp}

\subsection{Experimental Setup}
\label{sec:exp:setup}

\paragraph{Datasets.}
We evaluate on seven fine-grained visual recognition datasets: CUB-200-2011~\cite{wah2011cub} (200 bird species, 5{,}994 train / 5{,}794 test), Stanford Cars-196~\cite{krause2013cars} (196 car models), FGVC-Aircraft~\cite{maji2013aircraft} (100 aircraft variants), Oxford-IIIT Pets~\cite{parkhi2012pets} (37 pet breeds), Oxford Flowers-102~\cite{nilsback2008flowers} (102 flower categories), Describable Textures (DTD)~\cite{cimpoi2014dtd} (47 texture categories), and Food-101~\cite{bossard2014food101}.

For cross-domain experiments (\cref{sec:exp:component}), the SAE is trained on ImageNet-1K~\cite{deng2009imagenet} and applied zero-shot to all seven target datasets.

\paragraph{Metrics.}
We report Recall@K (R@K) for $K \in \{1, 5, 10, 20, 50, 100, 200\}$.
For the two-stage pipeline evaluation, the primary efficiency-relevant metric is
R@100 (first-stage recall coverage); for final ranking accuracy, we report R@1.
Query images are excluded from the reference set.

\paragraph{Implementation.}
All experiments use SigLIP2-SO400M~\cite{tschannen2025siglip2} as the frozen backbone.
The SAE uses expansion factor $e{=}16$ and SAE sparsity $k{=}16$ (non-zero activations per patch), extracting features from layer 26 (last transformer layer), with post-pool top-$k_{\text{post}}{=}16$ filtering and uint16 quantization applied at inference.
BM25 parameters are set to $k_1{=}1.5$, $b{=}0.75$ (standard values).
Feature extraction uses the frozen SigLIP2 backbone; BM25 scoring uses sparse matrix operations.
Unless otherwise noted, all BM25-V results below are \emph{cross-domain}: the SAE is trained on ImageNet-1K and applied zero-shot to each target dataset (primary config: L26/$e{=}16$/$k{=}16$/pptk+quant).

\paragraph{Compared systems.}
\begin{itemize}
  \item \textbf{BM25-V} (ours, sparse only): BM25 scoring on SAE visual words; sparse channel in isolation.
  \item \textbf{Dense}: cosine similarity on SigLIP2 MAP embeddings; dense channel in isolation.
  \item \textbf{FAISS-HNSW}~\cite{malkov2018efficient}: ANN search on dense MAP embeddings, HNSW32, $\text{ef}{=}100$.
  \item \textbf{FAISS-IVF+PQ}~\cite{johnson2021billion}: ANN with product quantization on dense MAP embeddings, $m{=}48$, adaptive nbits (4--8 bit), $n_\text{probe}{=}1$.
  \item \textbf{BM25-V + Dense rerank} (ours, two-stage): BM25-V top-$K$ candidate retrieval followed by dense cosine reranking (\cref{sec:method:twostage}).
\end{itemize}

% -----------------------------------------------------------------------
\subsection{Main Result: Two-Stage Retrieval}
\label{sec:exp:main}

\cref{tab:main} compares the two-stage pipeline against dense retrieval and ANN baselines.
The two-stage system (BM25-V top-200 $\to$ dense rerank) achieves near-exact-dense accuracy
across all benchmarks while enabling dramatic efficiency gains (\cref{sec:exp:efficiency}).
On the most challenging dataset (CUB-200), two-stage R@1 is within $1.2$\% of full dense retrieval;
on DTD and Flowers-102, it \emph{exceeds} full dense retrieval by $+0.7$\% and $+0.1$\% respectively,
as BM25-V's IDF-weighted sparse scores resolve fine-grained ambiguities that global cosine
similarity misses.
Averaged across seven datasets, the two-stage system achieves R@1 $= 0.857$, matching
full dense retrieval ($0.859$) within rounding ($-0.2$\%).
The first-stage BM25-V alone achieves R@100 $\geq 0.984$ and R@200 $\geq 0.993$ across all datasets, confirming near-complete candidate recall; full per-dataset recall curves are in the supplementary material.
Note that HNSW is already near-lossless ($\leq 0.1$\% of Dense on all datasets); reranking HNSW candidates with exact dense cosine would therefore recover Dense itself, which is already reported in \cref{tab:main}.
The two-stage gain comes not from correcting approximation error but from combining a \emph{structurally different} sparse signal with dense reranking.
We further validate on the standard instance retrieval benchmarks ROxford5k and RParis6k~\cite{radenovic2018revisiting}: on RParis6k, the BM25-V+Dense ensemble \emph{surpasses} dense retrieval by $+0.9$\% mAP (Medium), confirming complementary signal even beyond fine-grained classification (details in supplementary material).

\begin{table*}[tb]
  \caption{R@1/5/10 on cross-domain fine-grained retrieval (7 datasets).
           SAE trained on ImageNet-1K, zero-shot.
           Two-stage: BM25-V top-$K$ candidates $\to$ dense cosine rerank.
           $\dagger$ Dense-only methods (no interpretability).}
  \label{tab:main}
  \centering
  \resizebox{\linewidth}{!}{%
  \begin{tabular}{@{}l ccc|ccc|ccc|ccc|ccc|ccc|ccc@{}}
    \toprule
    & \multicolumn{3}{c}{CUB-200}
    & \multicolumn{3}{c}{Cars-196}
    & \multicolumn{3}{c}{Aircraft}
    & \multicolumn{3}{c}{Pets}
    & \multicolumn{3}{c}{Flowers-102}
    & \multicolumn{3}{c}{DTD}
    & \multicolumn{3}{c}{Food-101} \\
    \cmidrule(lr){2-4}\cmidrule(lr){5-7}\cmidrule(lr){8-10}%
    \cmidrule(lr){11-13}\cmidrule(lr){14-16}\cmidrule(lr){17-19}\cmidrule(l){20-22}
    Method
      & R@1 & R@5 & R@10
      & R@1 & R@5 & R@10
      & R@1 & R@5 & R@10
      & R@1 & R@5 & R@10
      & R@1 & R@5 & R@10
      & R@1 & R@5 & R@10
      & R@1 & R@5 & R@10 \\
    \midrule
    Dense$\dagger$
      & .767 & .948 & .975
      & .922 & .988 & .996
      & .707 & .915 & .966
      & .912 & .982 & .990
      & .989 & .997 & .998
      & .762 & .912 & .946
      & .954 & .987 & .992 \\
    FAISS-HNSW$\dagger$
      & .768 & .948 & .975
      & .922 & .988 & .996
      & .708 & .915 & .966
      & .912 & .982 & .990
      & .989 & .997 & .998
      & .762 & .912 & .946
      & .954 & .987 & .992 \\
    FAISS-IVF+PQ$\dagger$
      & .734 & .918 & .947
      & .897 & .978 & .987
      & .715 & .918 & .958
      & .915 & .957 & .965
      & .941 & .956 & .958
      & .706 & .847 & .880
      & .941 & .974 & .980 \\
    \midrule
    BM25-V (ours)
      & .472 & .738 & .835
      & .715 & .900 & .946
      & .523 & .765 & .851
      & .771 & .923 & .962
      & .954 & .986 & .993
      & .747 & .888 & .930
      & .865 & .947 & .965 \\
    Two-stage $K{=}100$ (ours)
      & .744 & .929 & .960
      & .914 & .983 & .990
      & .701 & .908 & .960
      & .916 & .979 & .988
      & .990 & .997 & .998
      & .769 & .919 & .955
      & .946 & .980 & .986 \\
    Two-stage $K{=}200$ (ours)
      & .755 & .941 & .969
      & .918 & .985 & .993
      & .704 & .914 & .965
      & .911 & .984 & .991
      & .991 & .997 & .998
      & .769 & .917 & .955
      & .950 & .983 & .988 \\
    \bottomrule
  \end{tabular}}
\end{table*}

% -----------------------------------------------------------------------
\subsection{Cross-Domain Results}
\label{sec:exp:component}

BM25-V generalizes across domains without per-dataset fine-tuning, and retrieves high-recall candidates at low cost: the encoding requires only a single linear projection per patch ($\mathbf{W}_e \mathbf{z} + \mathbf{b}_e$, followed by ReLU, top-$k$, and sum-pool), and gallery scoring is a sparse matrix operation over $k_{\text{post}}{=}16$ active dimensions per image.
Despite this lightweight design, BM25-V R@100 reaches $0.984$--$0.999$ across all seven datasets (CUB-200: $0.984$; Flowers-102: $0.999$), providing a near-complete recall pool for the dense rerank stage.
On datasets where local discriminative patterns dominate, BM25-V R@1 approaches the dense baseline: DTD ($0.747$ vs.\ $0.762$) and Flowers-102 ($0.954$ vs.\ $0.989$).
Dense retrieval integrates global scene context; BM25-V amplifies rare local patches via IDF---structurally non-redundant signals from the same frozen backbone, motivating the two-stage design (\cref{sec:method:twostage}).
Full recall curves (R@1--R@200) for all seven benchmarks are in the supplementary material.

% -----------------------------------------------------------------------
\subsection{Ablation Studies}
\label{sec:exp:ablation}

\paragraph{Scoring function.}
Under the primary configuration (with post-pool top-$k_{\text{post}}$), BM25 and dot-product scoring on identical sparse vectors achieve comparable R@1 across all seven datasets ($\leq 2$\% difference on most benchmarks).
This is expected: top-$k_{\text{post}}$ hard-filters each image vector to only the $k_{\text{post}}{=}16$ most-accumulated visual words, independently suppressing the pervasive dimensions that IDF would down-weight.
We adopt BM25 because it is the principled scoring function under the Zipfian distributional regime (\cref{sec:zipfian}): IDF provides continuous, magnitude-aware suppression of uninformative dimensions and naturally integrates with inverted-index infrastructure (posting lists, WAND pruning~\cite{broder2003efficient}).

\paragraph{Effect of sparsity $k$.}
\cref{tab:k} shows R@1 as a function of $k$ on CUB-200 (layer~$-1$, $e{=}4$, BM25).

\begin{table}[tb]
  \caption{Effect of sparsity $k$ on CUB-200-2011 (R@1 and sparse index size).
           Low $k$ preserves discriminative structure; $k{=}128$ collapses.}
  \label{tab:k}
  \centering
  \small
  \begin{tabular}{@{}lcc@{}}
    \toprule
    $k$ & R@1 & Index size (MB) \\
    \midrule
    16  & 0.508 & 35.8 \\
    32  & 0.483 & 61.7 \\
    64  & 0.396 & 97.2 \\
    128 & 0.038 & 140.6 \\
    \bottomrule
  \end{tabular}
\end{table}

At $k{=}128$, performance collapses to near-random ($\text{R@1}{=}0.038$).
This collapse is \emph{theoretically predicted}, not an unexpected failure: when too many visual words are active per patch, the sum-pooled image vector approaches dense, document frequencies converge ($\text{df}_i \to N$ for most dimensions, $\text{IDF}_i \to 0$), and the Zipfian discriminative structure that IDF relies on is destroyed.
This is the same phenomenon as dense bag-of-words without IDF in text retrieval.
The result confirms that maintaining the Zipfian regime (i.e., $k$ small enough that most SAE dimensions remain rare across the reference set) is a prerequisite for IDF-based scoring---and that $k \leq 32$ satisfies this condition across our benchmarks.

Ablations of expansion factor $e$ and backbone layer are provided in the supplementary material.
Briefly: larger $e$ consistently improves R@1 (we adopt $e{=}16$), and the final transformer layer is most discriminative (we use layer~26).

% -----------------------------------------------------------------------
\subsection{Efficiency Analysis}
\label{sec:exp:efficiency}

All efficiency numbers in this section are derived from the primary configuration:
SAE checkpoint \texttt{L26\_x16\_k16} (layer~26, expansion factor $e{=}16$,
top-$k$ sparsity $k{=}16$), with post-pool top-$k$ and uint16 quantization.
\textbf{Note on feature extraction cost.}
The ViT backbone forward pass and SAE encoding are \emph{shared} between BM25-V and the
dense baseline: both require the same SigLIP2 forward pass to obtain patch features
(BM25-V) and the MAP pooler embedding (dense rerank).
This shared cost is therefore not an overhead specific to BM25-V and is excluded from
the query-ops comparison below, which focuses on the \emph{scoring and retrieval stage}.
This gives a sparse vocabulary of $D_s = eD$ dimensions,
and post-pool top-$k_{\text{post}}$ (\cref{eq:pptk}) enforces exactly $L_0 = k_{\text{post}}$ non-zero
entries per image vector.

This extreme sparsity has direct efficiency consequences: the BM25-V index stores
only $L_0$ (index, value) pairs per image, and query scoring touches only the posting
lists of $L_0$ dimensions.
\cref{tab:efficiency} compares the resulting cost profile against standard baselines.

\begin{table}[tb]
  \caption{Efficiency comparison.
           Memory is per-vector storage; query cost counts scoring operations
           for one query against $N$ reference images.
           $D$: dense embedding dimension; $M$, ef: HNSW graph degree and search budget;
           $m$, $b_\text{PQ}$: PQ sub-quantizers and bytes per code;
           $n_p$, $n_l$: IVF probe and list counts;
           $L_0$: non-zeros per sparse vector;
           $C$: FLOPs per BM25 posting entry.
           Our config: $D{=}1152$, $M{=}32$, $\text{ef}{=}100$,
           $m{=}48$, $b_\text{PQ}{=}1$, $n_l{=}100$, $n_p{=}1$, $L_0{=}16$, $K{=}200$. \textsuperscript{$\dagger$}$V_\text{active}$ ranges from 527 to 3{,}240 across our benchmarks; as $N$ grows,
  $V_\text{active} \to D_s$ and cost $\to C L_0^2 N / D_s$ (\cref{eq:vactive}).}
  \label{tab:efficiency}
  \centering
  \small
  \setlength{\tabcolsep}{4pt}
  \begin{tabular}{@{}llrl@{}}
    \toprule
    Method & Memory/vec & Compression & Query ops \\
    \midrule
    Dense (exact)         & $4D$            & $1\times$            & $2DN$              \\
    HNSW~\cite{malkov2018efficient}
                          & $4(D{+}2M)$    & ${\approx}\,1\times$ & $2D \cdot \text{ef}$ \\
    IVF+PQ~\cite{jegou2011product}
                          & $m \!\cdot\! b_\text{PQ}$ & up to $192\times$ & $\tfrac{n_p}{n_l}\,m\,N$ \\
    \midrule
    BM25-V (ours)         & $6\,L_0$       & $48\times$           & $\tfrac{C \cdot L_0^2}{V_\text{active}}\,N$\textsuperscript{$\dagger$} \\
    Two-stage (ours)      & $4D{+}6\,L_0$  & ${\approx}\,1\times$ & $\tfrac{C \cdot L_0^2}{V_\text{active}}\,N {+} KD$ \\
    \bottomrule
  \end{tabular}

  \vspace{2pt}
\end{table}

\paragraph{Index memory.}
Let $L_0 = \|\mathbf{v}\|_0$ denote the number of non-zero entries per image vector
(exactly $k_{\text{post}}$ after post-pool top-$k_{\text{post}}$).
Each BM25-V vector stores $L_0$ entries, each comprising a 4-byte dimension
index and a 2-byte \texttt{uint16} activation value: $6\,L_0$~bytes/vec.
This is a $4D / 6\,L_0 = 48\times$ compression relative to the $4D$-byte float32 dense
embedding---comparable to IVF+PQ (up to $192\times$), but without quantization error.
In the two-stage pipeline, the sparse index adds only $6\,L_0$~bytes per image
on top of the dense embeddings that the reranker already requires,
bringing total two-stage memory to $(4D + 6\,L_0)$~bytes/vec.
By contrast, PQ achieves higher compression but at a recall cost (1--6\%; \cref{tab:main}).

\paragraph{Query cost.}
At query time, BM25-V does \emph{not} iterate over all $N$ reference-set images.
Instead, it traverses the inverted posting lists of only the $L_0$ active query dimensions---directly reading the small set of reference-set images that share at least one visual word with the query.
Each posting list has expected length $\overline{\text{df}} = N \cdot L_0 / V_\text{active}$,
where $V_\text{active}$ is the number of SAE dimensions that fire at least once in the reference set.
At benchmark scale, $V_\text{active} \ll D_s$: the Zipfian analysis (\cref{sec:zipfian}) finds $V_\text{active} \in [527, 3{,}240]$ across our seven datasets.
The expected number of scoring operations per query is therefore:
\begin{equation}
  L_0 \times \frac{N \cdot L_0}{V_\text{active}} \times C_\text{bm25}
  \;=\; \frac{C_\text{bm25} \cdot L_0^2}{V_\text{active}}\,N.
  \label{eq:bm25cost}
\end{equation}
Concretely, this yields 1{,}000--30{,}000 operations per query across our benchmarks
(\eg DTD: 1{,}002; CUB-200: 5{,}120; Food-101: 29{,}920).
These are the costs of the \emph{BM25-V first stage alone}.
For the full two-stage pipeline, the dense rerank of $K$ candidates adds
$KD$ operations---comparable to HNSW's constant cost of $2D \cdot \text{ef}$.
The two-stage therefore does not claim a query-latency advantage over HNSW;
its efficiency gain is specifically the replacement of an $O(ND)$
full dense scan with
sparse BM25-V scoring plus a small fixed dense rerank.
The BM25-V sparse operations are comparable to or below IVF+PQ's
$\tfrac{n_p}{n_l}\,m\,N$~\cite{jegou2011product}.

A natural question is whether the $V_\text{active}$ denominator approaches $D_s$
at larger reference set sizes.
Treating each image's $L_0$ active dimensions as independent draws from $D_s$,
the expected number of active dimensions after $N$ images follows the coupon-collector model:
\begin{equation}
  V_\text{active}(N) \;=\; D_s\!\left(1 - \left(1 - \frac{L_0}{D_s}\right)^{\!N}\right).
  \label{eq:vactive}
\end{equation}
As $N$ grows, $V_\text{active} \to D_s$ and the per-query cost converges to
$C_\text{bm25} \cdot L_0^2 \cdot N / D_s$---the formula one
obtains by assuming uniform spreading across all dimensions.
At our benchmark scales ($N \leq 76\text{K}$), this limit has not been reached,
so the corrected formula with $V_\text{active}$ gives the true cost.

Projected costs at industrial scale ($N{=}100\text{M}$) are provided in the supplementary material.
Briefly: at $N{=}100\text{M}$, BM25-V saturates its vocabulary ($V_\text{active} \to D_s$) and query cost
converges to $C_\text{bm25} L_0^2 N / D_s$---comparable to IVF+PQ in total ops.
The sparse index adds only $6\,L_0$ bytes per image on top of the dense vectors
that the two-stage reranker already requires.

\paragraph{Build time, updates, and scalability.}
Beyond query cost, the inverted-index structure offers several practical advantages
over graph- and quantization-based indices.
\emph{Build time:} HNSW construction is $O(N \log N \cdot M \cdot D)$---hours at
million scale. BM25-V index build is $O(N \cdot L_0)$: scatter $L_0$ entries per
image into posting lists, completing in seconds.
\emph{Dynamic updates:} inserting or deleting an image touches only $L_0$ posting
lists plus $L_0$ IDF counter updates---$O(L_0)$ with no structural degradation.
HNSW supports only tombstone deletion (graph quality degrades over time, requiring
periodic rebuilds), and IVF+PQ centroids grow stale as the data distribution shifts.
\emph{Sharding:} posting lists partition naturally by term ID across machines,
following the same architecture as web-scale text search engines.
HNSW's graph traversal requires cross-shard hops, making distributed deployment
harder.
Finally, BM25-V retrieves from a \emph{different representation space} than
dense methods: even when HNSW is faster at large $N$, combining the two
signals in a hybrid improves recall beyond what either achieves alone
(\cref{tab:main}).

\paragraph{Path to sub-linear query time.}
Integrating WAND pruning~\cite{broder2003efficient} would reduce BM25-V query cost
from $O(N)$ to empirical $O(\sqrt{N})$ while returning \emph{exact} top-$K$ results.
A detailed FLOP-level comparison with HNSW is provided in the supplementary material.

\paragraph{Empirical latency.}
Wall-clock benchmarks on CPU (Apple M4, $N$ up to 1M; supplementary material)
confirm the theoretical analysis: at $N{=}1\text{M}$, BM25-V query latency is
$5.2\times$ lower than dense exact search, and the two-stage pipeline is $3.5\times$
faster while maintaining near-exact accuracy.
Index build time is ${\approx}50{,}000\times$ faster than HNSW at $N{=}1\text{M}$
($0.09$\,s vs.\ $75$\,min), enabling real-time index updates.
GPU sparse scoring via cuSPARSE further closes the gap to $9\times$ of dense GPU
matmul at $N{=}1\text{M}$ (supplementary material).

%% file: chapters/zipfian.tex
\section{Visual Word Frequency Distribution}
\label{sec:zipfian}

\begin{figure}[tb]
  \centering
  \includegraphics[width=0.9\linewidth]{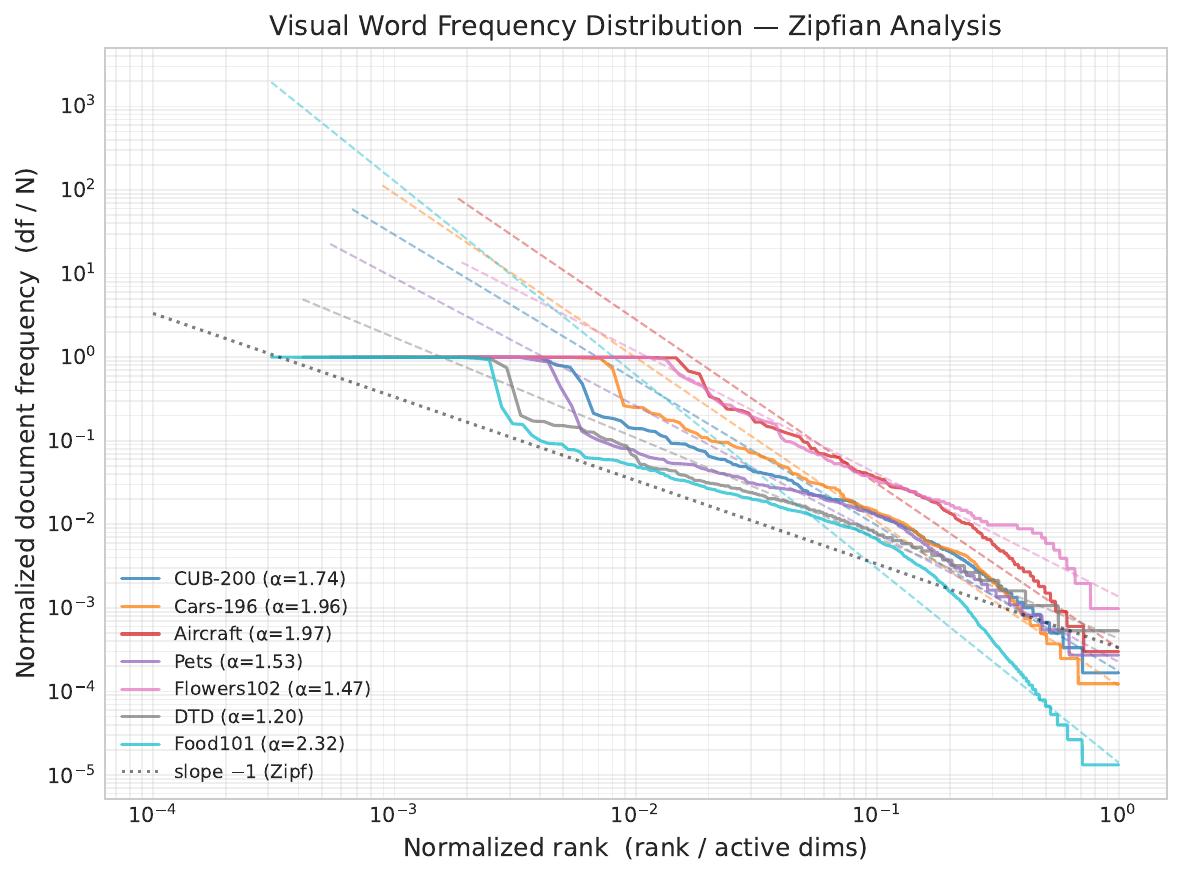}
  \caption{Normalized rank-frequency log-log plot for all 7 datasets.
           Each curve is normalized so rank-1 frequency $= 1$.
           Near-parallel lines confirm a common power-law family with
           $\alpha \in [1.20, 2.32]$; slopes steeper than text Zipf ($\alpha \approx 1$)
           show that visual words are more heavy-tailed than natural language.}
  \label{fig:zipfian}
\end{figure}

A central claim of BM25-V is that IDF weighting is \emph{principled}, not heuristic.
In text retrieval, IDF is justified by Zipf's law: word frequencies follow a power-law
distribution, so a small fraction of words are pervasive (and uninformative) while the
vast majority are rare (and discriminative)~\cite{zipf1949human}.
We verify that the same distributional structure holds for SAE-encoded visual words.

\paragraph{Measurement.}
For each dataset we compute the document frequency $\mathrm{df}_i$ of every active SAE
dimension across the reference set, sort by rank, and fit a power-law in log-log space.
Head dimensions fire in ${>}50\%$ of images; tail dimensions fire in ${<}10\%$.
All experiments use the primary config (layer~26, $e{=}16$, $k{=}16$, pptk+quant).

The power-law fit achieves $R^2 \in [0.917, 0.973]$ across all 7 datasets,
confirming genuine heavy-tailed structure across textures, birds, cars, food, and flowers---not
a dataset-specific artifact.
Fitted exponents $\alpha \in [1.20, 2.32]$ consistently exceed the text baseline ($\alpha \approx 1$):
visual words are \emph{more} heavy-tailed than language.
The head fraction---visual ``stop words'' activating in ${>}50\%$ of images---is tiny:
only $0.25\%$--$1.83\%$ of active dimensions.
Correspondingly, $96\%$--$99.7\%$ of active dimensions are discriminative (IDF ${>}2.0$),
so IDF amplifies virtually the entire active vocabulary rather than a small subset.
Per-dataset statistics are provided in the supplementary material.

This Zipfian structure directly motivates pervasive dimension suppression:
the head dimensions (${\approx}0.80\%$ of active dims) appear in most images and would dominate scores without down-weighting.
BM25's IDF assigns near-zero weight to these dimensions ($\mathrm{IDF}_i \approx 0$ when $\mathrm{df}_i \approx N$), while post-pool top-$k_{\text{post}}$ hard-filters to the most-accumulated rare dimensions---two complementary mechanisms grounded in the same distributional structure.

%% file: chapters/conclusion.tex
\section{Conclusion}
\label{sec:conclusion}

We presented \textbf{BM25-V}, a sparse visual retrieval module that applies Okapi BM25 scoring to Sparse Autoencoder features derived from ViT patch tokens.
The core insight is that SAE activations on deep visual features follow a Zipfian-like distribution, making IDF weighting the principled---not heuristic---scoring function for sparse visual retrieval.
Combined with post-pool top-$k_{\text{post}}$ filtering, which provides complementary hard suppression of pervasive dimensions, BM25-V enables cheap, high-recall candidate retrieval via sparse inverted-index operations.

The two-stage BM25-V pipeline combines local discriminative part patterns (sparse) with global class semantics (dense) from a shared frozen backbone.
These two signals are architecturally non-redundant, and their combination matches or exceeds full dense retrieval on all fine-grained benchmarks (avg $-0.2$~pp; DTD $+0.7$~pp; Flowers-102 $+0.1$~pp).
Unlike FAISS-PQ, which sacrifices accuracy for memory, BM25-V + Dense occupies a different operating point: near-exact accuracy with interpretability, at minimal overhead---the sparse index adds only 96 bytes per image ($48{\times}$ compression) and $(0.4\text{--}2.4)\,N$ sparse scoring operations per query at benchmark scale, replacing the $O(N{\cdot}D)$ full dense scan with a sparse first stage plus a small fixed dense rerank (\cref{sec:exp:efficiency}).

BM25-V is interpretable by construction: every retrieval decision is attributable to named visual words with quantified IDF scores.
To our knowledge, BM25-V is among the first to provide quantified, IDF-weighted token-level attribution for dense-backbone image retrieval.

% \paragraph{Limitations.}
% BM25-V requires patch features from the deep backbone layers, coupling it to the same forward pass as dense retrieval.
% SAE training adds a preprocessing step per dataset (or per domain for zero-shot use).
% The efficiency analysis relies on theoretical projections derived from the coupon-collector model (\cref{eq:vactive}); empirical validation at genuinely large scale ($N \gg 100$K, e.g., ROxford, GLDv2) remains future work.
% The $k{=}128$ collapse is theoretically predicted (IDF scores collapse when sparsity is insufficient to maintain the Zipfian regime) rather than an unexpected failure, but establishing a principled rule for choosing $k$ across arbitrary datasets warrants further study.

%% file: main.bib
@String(SIGIR = {ACM SIGIR})

@String(EMNLP = {Proc. Conf. Empirical Methods Natural Language Process.})

@ARTICLE{pq,
  author={Jégou, Herve and Douze, Matthijs and Schmid, Cordelia},
  journal={IEEE Transactions on Pattern Analysis and Machine Intelligence}, 
  title={Product Quantization for Nearest Neighbor Search}, 
  year={2011},
  volume={33},
  number={1},
  pages={117-128},
  doi={10.1109/TPAMI.2010.57}
}

@inbook{robertson1976relevance,
    author = {Robertson, Stephen E. and Sparck Jones, Karen},
    title = {Relevance weighting of search terms},
    year = {1988},
    isbn = {0947568212},
    publisher = {Taylor Graham Publishing},
    address = {GBR},
    booktitle = {Document Retrieval Systems},
    pages = {143–160},
    numpages = {18}
}

@inproceedings{robertson1994okapi,
  title={Okapi at TREC-3},
  author={Stephen E. Robertson and Steve Walker and Susan Jones and Micheline Hancock-Beaulieu and Mike Gatford},
  booktitle={Text Retrieval Conference},
  year={1994},
  url={https://api.semanticscholar.org/CorpusID:41563977}
}

@inproceedings{formal2021splade,
    author = {Formal, Thibault and Piwowarski, Benjamin and Clinchant, St\'{e}phane},
    title = {SPLADE: Sparse Lexical and Expansion Model for First Stage Ranking},
    year = {2021},
    isbn = {9781450380379},
    publisher = {Association for Computing Machinery},
    address = {New York, NY, USA},
    url = {https://doi.org/10.1145/3404835.3463098},
    doi = {10.1145/3404835.3463098},
    booktitle = {Proceedings of the 44th International ACM SIGIR Conference on Research and Development in Information Retrieval},
    pages = {2288–2292},
    numpages = {5},
    location = {Virtual Event, Canada},
    series = {SIGIR '21}
}

@article{dai2020deepct,
  title={Context-aware sentence/passage term importance estimation for first stage retrieval},
  author={Dai, Zhuyun and Callan, Jamie},
  journal={arXiv preprint arXiv:1910.10687},
  year={2019}
}

@article{nogueira2019doc2query,
  title={Document expansion by query prediction},
  author={Nogueira, Rodrigo and Yang, Wei and Lin, Jimmy and Cho, Kyunghyun},
  journal={arXiv preprint arXiv:1904.08375},
  year={2019}
}

@article{lin2021few,
  title={A few brief notes on deepimpact, coil, and a conceptual framework for information retrieval techniques},
  author={Lin, Jimmy and Ma, Xueguang},
  journal={arXiv preprint arXiv:2106.14807},
  year={2021}
}

@inproceedings{karpukhin2020dense,
  title={Dense passage retrieval for open-domain question answering},
  author={Karpukhin, Vladimir and Oguz, Barlas and Min, Sewon and Lewis, Patrick and Wu, Ledell and Edunov, Sergey and Chen, Danqi and Yih, Wen-tau},
  booktitle={Proceedings of the 2020 conference on empirical methods in natural language processing (EMNLP)},
  pages={6769--6781},
  year={2020}
}

@article{bricken2023monosemanticity,
       title={Towards Monosemanticity: Decomposing Language Models With Dictionary Learning},
       author={Bricken, Trenton and Templeton, Adly and Batson, Joshua and Chen, Brian and Jermyn, Adam and Conerly, Tom and Turner, Nick and Anil, Cem and Denison, Carson and Askell, Amanda and Lasenby, Robert and Wu, Yifan and Kravec, Shauna and Schiefer, Nicholas and Maxwell, Tim and Joseph, Nicholas and Hatfield-Dodds, Zac and Tamkin, Alex and Nguyen, Karina and McLean, Brayden and Burke, Josiah E and Hume, Tristan and Carter, Shan and Henighan, Tom and Olah, Christopher},
       year={2023},
       journal={Transformer Circuits Thread},
       note={https://transformer-circuits.pub/2023/monosemantic-features/index.html}
}

@inproceedings{park2025decoding,
  title={Decoding dense embeddings: Sparse autoencoders for interpreting and discretizing dense retrieval},
  author={Park, Seongwan and Kim, Taeklim and Ko, Youngjoong},
  booktitle={Proceedings of the 2025 Conference on Empirical Methods in Natural Language Processing},
  pages={26479--26496},
  year={2025}
}

@inproceedings{formal2026learning,
    title={Learning Retrieval Models with Sparse Autoencoders},
    author={Thibault Formal and Maxime Louis and Herv{\'e} D{\'e}jean and St{\'e}phane Clinchant},
    booktitle={The Fourteenth International Conference on Learning Representations},
    year={2026},
    url={https://openreview.net/forum?id=TuFjICawSc}
}

@inproceedings{radford2021learning,
  title={Learning transferable visual models from natural language supervision},
  author={Radford, Alec and Kim, Jong Wook and Hallacy, Chris and Ramesh, Aditya and Goh, Gabriel and Agarwal, Sandhini and Sastry, Girish and Askell, Amanda and Mishkin, Pamela and Clark, Jack and others},
  booktitle={International conference on machine learning},
  pages={8748--8763},
  year={2021},
  organization={PmLR}
}

@article{oquab2023dinov2,
  title={Dinov2: Learning robust visual features without supervision},
  author={Oquab, Maxime and Darcet, Timoth{\'e}e and Moutakanni, Th{\'e}o and Vo, Huy and Szafraniec, Marc and Khalidov, Vasil and Fernandez, Pierre and Haziza, Daniel and Massa, Francisco and El-Nouby, Alaaeldin and others},
  journal={arXiv preprint arXiv:2304.07193},
  year={2023}
}

@inproceedings{zhai2023sigmoid,
  title={Sigmoid loss for language image pre-training},
  author={Zhai, Xiaohua and Mustafa, Basil and Kolesnikov, Alexander and Beyer, Lucas},
  booktitle={Proceedings of the IEEE/CVF international conference on computer vision},
  pages={11975--11986},
  year={2023}
}

@article{tschannen2025siglip2,
  title={Siglip 2: Multilingual vision-language encoders with improved semantic understanding, localization, and dense features},
  author={Tschannen, Michael and Gritsenko, Alexey and Wang, Xiao and Naeem, Muhammad Ferjad and Alabdulmohsin, Ibrahim and Parthasarathy, Nikhil and Evans, Talfan and Beyer, Lucas and Xia, Ye and Mustafa, Basil and others},
  journal={arXiv preprint arXiv:2502.14786},
  year={2025}
}

@article{johnson2021billion,
  title={Billion-scale similarity search with GPUs},
  author={Johnson, Jeff and Douze, Matthijs and J{\'e}gou, Herv{\'e}},
  journal={IEEE transactions on big data},
  volume={7},
  number={3},
  pages={535--547},
  year={2019},
  publisher={IEEE}
}

@article{malkov2018efficient,
  title={Efficient and robust approximate nearest neighbor search using hierarchical navigable small world graphs},
  author={Malkov, Yu A and Yashunin, Dmitry A},
  journal={IEEE transactions on pattern analysis and machine intelligence},
  volume={42},
  number={4},
  pages={824--836},
  year={2018},
  publisher={IEEE}
}

@inproceedings{sivic2003video,
  author={Sivic and Zisserman},
  booktitle={Proceedings Ninth IEEE International Conference on Computer Vision}, 
  title={Video Google: a text retrieval approach to object matching in videos}, 
  year={2003},
  volume={},
  number={},
  pages={1470-1477 vol.2},
  doi={10.1109/ICCV.2003.1238663}}

@inproceedings{jegou2010aggregating,
  author={Jégou, Hervé and Douze, Matthijs and Schmid, Cordelia and Pérez, Patrick},
  booktitle={2010 IEEE Computer Society Conference on Computer Vision and Pattern Recognition}, 
  title={Aggregating local descriptors into a compact image representation}, 
  year={2010},
  volume={},
  number={},
  pages={3304-3311},
  doi={10.1109/CVPR.2010.5540039}}

@article{jegou2011product,
  author={Jégou, Herve and Douze, Matthijs and Schmid, Cordelia},
  journal={IEEE Transactions on Pattern Analysis and Machine Intelligence}, 
  title={Product Quantization for Nearest Neighbor Search}, 
  year={2011},
  volume={33},
  number={1},
  pages={117-128},
  doi={10.1109/TPAMI.2010.57}
  }

@inproceedings{krause2013cars,
  author={Krause, Jonathan and Stark, Michael and Deng, Jia and Fei-Fei, Li},
  booktitle={2013 IEEE International Conference on Computer Vision Workshops}, 
  title={3D Object Representations for Fine-Grained Categorization}, 
  year={2013},
  volume={},
  number={},
  pages={554-561},
  doi={10.1109/ICCVW.2013.77}
}

@article{maji2013aircraft,
  title={Fine-grained visual classification of aircraft},
  author={Maji, Subhransu and Rahtu, Esa and Kannala, Juho and Blaschko, Matthew and Vedaldi, Andrea},
  journal={arXiv preprint arXiv:1306.5151},
  year={2013}
}

@inproceedings{parkhi2012pets,
  author={Parkhi, Omkar M and Vedaldi, Andrea and Zisserman, Andrew and Jawahar, C. V.},
  booktitle={2012 IEEE Conference on Computer Vision and Pattern Recognition}, 
  title={Cats and dogs}, 
  year={2012},
  volume={},
  number={},
  pages={3498-3505},
  doi={10.1109/CVPR.2012.6248092}
}

@inproceedings{nilsback2008flowers,
  author={Nilsback, Maria-Elena and Zisserman, Andrew},
  booktitle={2008 Sixth Indian Conference on Computer Vision, Graphics \& Image Processing}, 
  title={Automated Flower Classification over a Large Number of Classes}, 
  year={2008},
  volume={},
  number={},
  pages={722-729},
  doi={10.1109/ICVGIP.2008.47}
}

@inproceedings{cimpoi2014dtd,
  title={Describing textures in the wild},
  author={Cimpoi, Mircea and Maji, Subhransu and Kokkinos, Iasonas and Mohamed, Sammy and Vedaldi, Andrea},
  booktitle={Proceedings of the IEEE conference on computer vision and pattern recognition},
  pages={3606--3613},
  year={2014}
}

@inproceedings{bossard2014food101,
  title = {Food-101 -- Mining Discriminative Components with Random Forests},
  author = {Bossard, Lukas and Guillaumin, Matthieu and Van Gool, Luc},
  booktitle = {European Conference on Computer Vision},
  year = {2014}
}

@inproceedings{deng2009imagenet,
  author={Deng, Jia and Dong, Wei and Socher, Richard and Li, Li-Jia and Kai Li and Li Fei-Fei},
  booktitle={2009 IEEE Conference on Computer Vision and Pattern Recognition}, 
  title={ImageNet: A large-scale hierarchical image database}, 
  year={2009},
  volume={},
  number={},
  pages={248-255},
  doi={10.1109/CVPR.2009.5206848}
}

@inproceedings{broder2003efficient,
    author = {Broder, Andrei Z. and Carmel, David and Herscovici, Michael and Soffer, Aya and Zien, Jason},
    title = {Efficient query evaluation using a two-level retrieval process},
    year = {2003},
    isbn = {1581137230},
    publisher = {Association for Computing Machinery},
    address = {New York, NY, USA},
    url = {https://doi.org/10.1145/956863.956944},
    doi = {10.1145/956863.956944},
    booktitle = {Proceedings of the Twelfth International Conference on Information and Knowledge Management},
    pages = {426–434},
    numpages = {9},
    location = {New Orleans, LA, USA},
    series = {CIKM '03}
}

@book{zipf1949human,
  title={Human Behavior and the Principle of Least Effort: An Introduction to Human Ecology},
  author={Zipf, George Kingsley},
  year={1949},
  publisher={Addison-Wesley}
}

@inproceedings{radenovic2018revisiting,
  title={Revisiting oxford and paris: Large-scale image retrieval benchmarking},
  author={Radenovi{\'c}, Filip and Iscen, Ahmet and Tolias, Giorgos and Avrithis, Yannis and Chum, Ond{\v{r}}ej},
  booktitle={Proceedings of the IEEE conference on computer vision and pattern recognition},
  pages={5706--5715},
  year={2018}
}

@article{wah2011cub, 
    title={The Caltech-UCSD Birds-200-2011 Dataset},  
  publisher={California Institute of Technology}, 
  author={Wah, Catherine and Branson, Steve and Welinder, Peter and Perona, Pietro and Belongie, Serge}, year={2011}, 
  month={7} 
}
